\documentclass{llncs}
\usepackage{epsf}

\usepackage{tikz-dependency}
\usepackage{cleveref} 
\usepackage{subcaption} 
\usepackage{booktabs}
\usepackage{graphicx}
\usepackage{makecell}
\usepackage{multirow}
\usepackage{adjustbox}
\usepackage{dsfont}
\usepackage{fancyhdr}
\begin{document}

\title{Speeding up Natural Language Parsing by Reusing Partial Results}

\author{Michalina Strzyz \qquad Carlos G\'omez-Rodr\'iguez
  }
\institute{Universidade da Coru\~na, CITIC \\
  FASTPARSE Lab, LyS Research Group, Departamento de Computaci\'on \\
  Campus de Elvi\~na, s/n, 15071 A Coru\~na, Spain\\
  {\tt \{michalina.strzyz,carlos.gomez\}@udc.es}}

\maketitle

\begin{abstract}

This paper proposes a novel technique that applies case-based reasoning in order to generate templates for reusable parse tree fragments, based on PoS tags of bigrams and trigrams that demonstrate low variability in their syntactic analyses from prior data. The aim of this approach is to improve the speed of dependency parsers by avoiding redundant calculations. This can be resolved by applying the predefined templates that capture results of previous syntactic analyses and directly assigning the stored structure to a new n-gram that matches one of the templates, instead of parsing a similar text fragment again. The study shows that using a heuristic approach to select and reuse the partial results increases parsing speed by reducing the input length to be processed by a parser. The increase in parsing speed comes at some expense of accuracy. Experiments on English show promising results: the input dimension can be reduced by more than 20\% at the cost of less than 3 points of Unlabeled Attachment Score.

\end{abstract}

\section{Introduction}

Current state-of-art parsing algorithms are facing high computational costs, which can be an obstacle when applied to large-scale processing. For example, the BIST parser \cite{bistparser} reports speeds of around 50 sentences per second in modern CPUs, and even the fastest existing parsers, which forgo recurrent neural networks, are limited to a few hundred sentences per second \cite{udpipe:2017}. While these speeds can be acceptable for small-scale processing, they are clearly prohibitive when the intention is to apply natural language parsing at the web scale. Thus, there is a need for approaches that can parse faster, even if this comes at some accuracy cost, as differences in accuracy above a certain threshold have been shown to be unimportant for some downstream tasks \cite{imp}.

In this paper we propose a novel approach to improve the speed of existing parsers by avoiding redundant calculations.
In particular, we identify fragments of the input for which a syntactic parse is known with high confidence from prior data, so that we can reuse the existing result directly instead of parsing the fragment again. This effectively reduces the length of the input being processed by the parser, which is a major factor determining parsing time.

We test a prototype of the approach where the reusable fragments are bigrams and trigrams, the matching criteria are based on part-of-speech tags, and the parsers to be optimized are linear-time transition-based dependency parsers. Note, however, that the technique is generic enough to be applied to any kind of parsing algorithm (including constituent parsers) and speed improvements are expected to be higher for higher-complexity parsers such as those based on dynamic programming, which have been the traditional target of pruning and optimization techniques \cite{Bodenstab2011,VieiraEisner2017}.

The aim of this approach is to significantly improve the parsing speed while keeping an acceptable accuracy level.
This involves a trade-off between speed and accuracy, depending on how aggressively the technique is applied: more lenient confidence criteria for reusing fragments will lead to larger reductions in input length and thus faster parsing but at a cost to accuracy. We expect that the accuracy cost can be reduced in the future by using more fine-grained matching criteria (based on forms or lemmas) and augmenting training data so more fragments can be confidently reused.

\section{Reuse of Partial Results}

A natural language obeys the so-called Zipf's law, which shows how words are distributed with respect to their frequency. Namely, 
a language has a few high-frequency tokens and many uncommon tokens.
This phenomenon is also present when investigating lemmas \cite{baroni}. Similarly, n-gram phrases fit this distribution \cite{ha2002,ha2009}. Therefore, to some extent, repetitions of identical n-grams are likely to be found in large corpora.

This implies that a parser will probably encounter known n-grams in a text. Since most of the time repetitions of the same n-gram will have identical syntactic structure (e.g. the phrase ``the 10 provinces of Canada'' can be reasonably expected to always have the same internal structure), we can exploit this by reusing previous analyses for these known n-grams.
More generally, this can be extended to similar n-grams (``the 50 provinces of Spain'' can be expected to have the same analysis as the phrase above), which can be identified with templates: the two examples above can be captured by a template ``the CD provinces of NNP'' (where CD and NNP are the part-of-speech tags for numerals and proper nouns), associated with a partial syntactic analysis to be reused. This combination of template matching and result reuse can be seen as an instantiation of case-based reasoning, a cognitively-rooted problem-solving technique based on solving new problems by reusing solutions to similar past problems \cite{Richter2005,Hullermeier2007}.

Given a template like in the example above, we implement case-based reasoning by examining the training set and counting the number of n-grams that match it, as well as the different syntactic analyses that they have been assigned. If the variability of these analyses is higher than a given threshold, then the template will not be useful. However, if matching n-grams exhibit the same syntactic structure the vast majority of the time, then we can create a rule assigning that parse tree fragment to the template. At test time, we will reuse the parse tree fragment whenever we encounter an n-gram matching this rule.

\section{Generating the Templates}
In this prototype we will use limited training data. Since the frequency of word pairs is higher than multi-word phrases \cite{storing} only templates consisting of bigrams or trigrams of PoS tags are considered in this implementation in order to not suffer excessively from  sparsity problems. More detailed templates (such as those including lemmas or word forms) would require augmenting the training set.

In order to calculate the level of confidence of a syntactic analysis for a given template, the following patterns for a bigram of PoS tags, as detailed in \Cref{fig:dep}, are taken into consideration. 

\begin{figure}[] 
\begin{subfigure}{\columnwidth}
\centering
\begin{subfigure}[b]{0.5\columnwidth}
\begin{dependency}[theme = default]
   \begin{deptext}[column sep=1em] 
  
      NOUN \& VERB \&  \& Pattern:\\
    0 \& 1 \&  \& 1  - false\\
   \end{deptext}
   
   \depedge{2}{1}{nsubj} 
  \wordgroup{1}{1}{1}{noun}
 
\end{dependency}
\end{subfigure}
\begin{subfigure}[b]{0.5\columnwidth}
\begin{dependency}[theme = default]
   \begin{deptext}[column sep=1em]
      VERB \& NOUN \& \& Pattern:\\
       0 \& 1  \&  \& -  0 false\\
   \end{deptext}
   
   \depedge{1}{2}{obj}
   \wordgroup{1}{2}{2}{noun2}
 
\end{dependency}
\end{subfigure}
\caption{bigram with dependents with no descendants outside the scope}
\label{fig:bi1}
\end{subfigure}

\begin{subfigure}{\columnwidth}
\centering
\begin{subfigure}[b]{0.5\columnwidth}
\begin{dependency}[theme = default]
   \begin{deptext}[column sep=1em]
       \& NOUN  \& VERB  \& Pattern:\\
       \& 0 \& 1  \& 1  - true\\
   \end{deptext}
   
   \depedge[hide label, edge style={black,densely dotted}]{2}{1}{}
   \depedge{3}{2}{nsubj}
\end{dependency}
\end{subfigure}
\begin{subfigure}[b]{0.5\columnwidth}
\begin{dependency}[theme = default]
   \begin{deptext}[column sep=1em]
      VERB \& NOUN   \& \& Pattern:\\
        0 \& 1 \& \& - 0 true\\
   \end{deptext}
   
   \depedge[hide label, edge style={black,densely dotted}]{2}{3}{}
   \depedge{1}{2}{obj}
\end{dependency}
\end{subfigure}
\caption{bigram with a dependent pointing to a word outside the scope}
\label{fig:bi2}
\end{subfigure}

\begin{subfigure}{\columnwidth}
\centering
\begin{subfigure}[b]{0.5\columnwidth}
\begin{dependency}[label theme = default]
   \begin{deptext}[column sep=1em] 
  
      NOUN \& VERB  \& \& Pattern:\\
       0 \& 1\& \& - - false\\
   \end{deptext}
   
   \depedge[cross out]{2}{1}{}
\end{dependency}
\end{subfigure}\hspace{7mm}

\caption{bigram with missing relation}
\label{fig:bi3}
\end{subfigure}
\caption{Possible patterns for a bigram of PoS tags NOUN VERB and VERB NOUN used for calculating the confidence of a template. Each pattern denotes the index of each word's head ("-" in case of headless word) and whether a dependent is pointing to a word outside the scope of the bigram ("true") or 
not
("false").}
\label{fig:dep}
\end{figure}

To generate a template for a given PoS tag bigram, we count the frequency of each of these head patterns relative to the total frequency of the bigram in the training set. Then, we focus on the most frequent pattern for the bigram. If this pattern has a dependent pointing to a word outside the bigram (\Cref{fig:bi2}) or multiple unconnected roots (\Cref{fig:bi3}), then the bigram will be discarded for template generation. The reason is that our reuse approach is based on replacing a sentence fragment (for which the parse is extracted from a rule) with its head word. The parser will operate on this head word only, and the parsed fragment will then be linked with the resulting tree. To do this, reusable fragments must have a single head and no external material depending on their dependents.

If the dominant pattern is eligible according to these criteria, then we will consider it if its relative frequency (confidence) is above a given threshold (head threshold) and, since our parsing is labeled, if the most frequent dependency label involved in the pattern has, in turn, a relative frequency above a second threshold (label threshold).

The confidence of heads and labels for a trigram is calculated analogously. The number of possible patterns of a trigram increases to a total of 19. As can be seen in \Cref{fig:tri}, a trigram can be a fully connected tree where the dependents 
have no descendants outside the scope
(7 patterns) or where at least one of the dependents points to a word outside the scope (7 patterns) \footnote{In this implementation we only consider projective trees for trigrams.}. A trigram can also be not fully connected or not connected at all (5 patterns). 

\begin{figure}
\begin{subfigure}{\columnwidth}
\centering
\begin{subfigure}[b]{0.5\columnwidth} 
\begin{dependency}[theme = default]
   \begin{deptext}[column sep=1em] 
  
      NOUN \& ADV \& VERB \& \&   Pattern:\\
       0 \& 1 \& 2 \& \& 2 2 - false\\
   \end{deptext}
   
   \depedge{3}{1}{nsubj}
     \depedge{3}{2}{advmod}
  \wordgroup{1}{1}{1}{noun}
  \wordgroup{1}{2}{2}{adv}
 
\end{dependency}
\end{subfigure}
\caption{trigram with dependents 
with no descendants outside the scope
}
\label{fig:1}
\end{subfigure}

\begin{subfigure}{\columnwidth}
\centering
\begin{subfigure}[b]{0.5\columnwidth}
\begin{dependency}[theme = default]
   \begin{deptext}[column sep=1em]
     \&  NOUN \& ADV \& VERB  \&   Pattern:\\
      \& 0 \& 2 \& 2 \&  2 2 - true\\
   \end{deptext}
   
   \depedge[hide label, edge style={black,densely dotted}]{2}{1}{}
    \depedge{4}{2}{nsubj}
     \depedge{4}{3}{advmod}
  
   \wordgroup{1}{3}{3}{noun}
\end{dependency}
\end{subfigure}

\caption{trigram with a dependent pointing to a word outside the scope}
\label{fig:2}
\end{subfigure}

\begin{subfigure}{\columnwidth}
\centering
\begin{subfigure}[b]{0.45\columnwidth}
\begin{dependency}[label theme = default]
   \begin{deptext}[column sep=1em] 
  
     NOUN \& ADV \& VERB \& \& Pattern:\\
      0 \& 1 \& 2 \& \& - 2 - false\\
   \end{deptext}
   
   \depedge[cross out]{3}{1}{}
    \depedge{3}{2}{advmod}
     \wordgroup{1}{2}{2}{noun}
\end{dependency}
\end{subfigure}

\caption{trigram with missing relation with one of its components }
\label{fig:3}
\end{subfigure}
\caption{Some of the possible patterns for a trigram of PoS tags NOUN ADV VERB used for calculating the confidence of a template. Each pattern denotes the index of word's head ("-" in case of headless word) and whether a dependent is pointing to a word outside the scope of the trigram ("true") or 
not
("false"). }
\label{fig:tri}
\end{figure}

Similarly to the case of bigrams, only patterns for trigrams highlighted in \Cref{fig:1} that exceed predefined thresholds for confidence will be used as templates. If the input fragment matches a template, all 
dependents will be removed from the input to be processed by a parser. The patterns from \Cref{fig:2} and \Cref{fig:3} are discarded from being candidates for a template.

\section{Experiments}

\subsection{Data and Evaluation}\label{sec:data}
We conducted our experiments on the English treebank from Universal Dependencies (UD) v2.1 \cite{universal} and the results are evaluated with Unlabeled and Labeled Attachment Scores (UAS and LAS).  The speeds are measured in tokens/second on CPU \footnote{Intel Core i7-7700 CPU 4.2 GHz}.
\subsection{Model}\label{sec:model}

During training time our model computes a set of rules that surpasses two thresholds: one for the dominant heads and the second for the dominant labels for a given n-gram. In our experiments the thresholds were set manually. Each template contains information about the n-gram's most likely head(s) and label(s) in order to automatically assign a syntactic analysis to the input that matches that template 
at parsing time.

As an example, \Cref{tab:threshEx} illustrates some templates that surpass the thresholds in the training set. The thresholds were set to 83\% for confidence of head and 83\% for confidence of label. This generated 141 unique templates, of which 97 had confidence of 100\% for both thresholds but with low frequency.    

\renewcommand{\arraystretch}{1.2}
\begin{table*}[]
\centering
\caption{Example of templates generated during training time that surpass the predefined thresholds: 83\% for confidence of the dominant head pattern and 83\% for confidence of the dominant label pattern for a given n-gram of PoS tags. }
\begin{adjustbox}{max width=\textwidth}
\begin{tabular}{@{}cccccccc@{}}
\toprule
Template		& \multicolumn{1}{p{2cm}}{\centering Dominant head pattern}       & \multicolumn{1}{p{2cm}}{\centering Confidence of head pattern ($\%$)} & \multicolumn{1}{p{2cm}}{\centering Dominant label pattern} & \multicolumn{1}{p{2cm}}{\centering  Confidence of label pattern ($\%$)} \\ \midrule
DET NOUN    & 1 - false   & 86.50  & det - false    & 86.49   \\
SCONJ PROPN VERB   &     2 2 - false & 100 & mark nsubj - false & 100   \\
AUX ADJ       & 1 - false & 93.71 & cop - false & 93.64                \\
ADV VERB       & 1 - false & 93.17 & advmod - false & 91.31                                             \\ \bottomrule
\end{tabular}%
\end{adjustbox}
\label{tab:threshEx}
\end{table*}

The rules are applied to matching bigrams and trigrams on the training set, removing words other than the head.\footnote{In case a bigram and trigram overlap, the n-gram with higher head confidence will be chosen and its dependents will be removed.} This produces a reduced training set that no longer contains n-grams matching any of the templates. Our parser is then trained on this reduced training set. At parsing time templates are applied to the input, which is reduced in the same way, the parser is then ran on this shorter input and finally the parse tree fragments are attached to the resulting output. We verified experimentally that, as expected, a parser achieves better accuracy combined with our technique when trained on the reduced than on the entire training set.

\subsection{Results}\label{sec:results}

In our experiments we run the following dependency parsers on the remaining test set: transition-based BIST Parser \cite{bistparser} that uses bidirectional long short memory (BiLSTM) networks, as well as MaltParser \cite{nivre2006} with the arc-eager and Covington transition systems. 

\Cref{tab:thresholds} demonstrates the results after applying templates with threshold variation on the development set. In the experimental setup we use templates consisting of bigrams alone, trigrams alone or combined in order to compare their performance. While the specific parameters to choose depend on the speed-accuracy balance one wants to achieve, 
we selected the models where the thresholds for the dominant head and label pattern are 83-83 and 87-87 as our ``optimal models'', considering that they provide a reasonable trade-off between speed and accuracy.
Thus, only these optimal thresholds were used afterwards when testing the technique on the development and test sets in order to investigate more in detail and to compare the accuracy and parsing speed. We use the notation $M^{\mathrm{x,y}}_{\mathrm{z}}$ where $x$ indicates whether bigrams (2) were used, $y$ trigrams (3) and $z$ level of confidence for head and label pattern respectively.  

\Cref{tab:merged} compares the performance of the parsers with and without applying our technique. The results are revealing in several ways. The experiments confirm that the more lenient confidence thresholds result in larger reductions of input length and thus faster parsing, but at the expense of accuracy. This applies to all parsers and indicates that our approach can be generic and applicable to diverse parsing algorithms. 
Moreover, the results show that it is feasible to reduce input size by more than 20\% at the cost of less than 3 points of unlabeled attachment score.

\renewcommand{\arraystretch}{1.5}
\begin{table}[]
\centering
\caption{Performance of MaltParser with arc-eager transition system and the \% of the text reduction after applying templates with different thresholds, on the development set. The total number of words in the development set: 25150 and test set: 25097.}
\begin{adjustbox}{max width=\columnwidth}
\begin{tabular}{@{}cccccccc@{}}
\toprule
Setup	      & UAS ($\%$) & LAS ($\%$) & Word Reduction ($\%$) \\ \midrule
$M^{\mathrm{2,3}}_{\mathrm{90-90}}$         & 84.73 & 82.15                 & 5.0    \\
$M^{\mathrm{2,3}}_{\mathrm{87-87}}$           & \textbf{84.38} & \textbf{81.64}                  & \textbf{8.3}    \\
$M^{\mathrm{3}}_{\mathrm{85-85}}$   & 84.15 & 81.45                 & 8.5   \\
$M^{\mathrm{2}}_{\mathrm{85-85}}$         & 84.01 & 81.17                 & 11.1   \\
$M^{\mathrm{2,3}}_{\mathrm{85-85}}$           & 83.1  & 80.15                 & 18.2 \\
$M^{\mathrm{2,3}}_{\mathrm{83-83}}$      & \textbf{82.95} & \textbf{80.03}                 & \textbf{20.7}   \\
$M^{\mathrm{2,3}}_{\mathrm{80-80}}$           & 81.51 & 78.42                 &22.7   \\
$M^{\mathrm{2,3}}_{\mathrm{80-70}}$       & 81.2 & 77.71                 & 24.2   \\
 \bottomrule
\end{tabular}%
\end{adjustbox}
\label{tab:thresholds}
\end{table}

\renewcommand{\arraystretch}{1.2}
\begin{table*}[]
\centering
\caption{Performance of BIST Parser and MaltParser with the arc-eager and Covington transition systems and after applying templates compared with the baseline. The reported parsing speed (tokens/sec) only refers to the runtime of the dependency parser on the entire data set (baseline) or remaining text (that was passed to the parser after extracting the fragments captured by the templates) excluding the time needed to run the technique which is already negligible and will be optimized in the future versions.}

\begin{adjustbox}{max width=\textwidth}
\begin{tabular}{@{}ccccccccc@{}}
\toprule
Parser & Data Set  & Setup & UAS ($\%$) & LAS ($\%$) & Word Reduction ($\%$) & Tokens/Sec & Speed-up Factor \\ \midrule
\multirow{7}{*}{BIST Parser} & \multirow{3}{*}{dev}  & \textbf{baseline}                                                                                             & \textbf{88.07}      & \textbf{86.08}      & \textbf{NA}  & \textbf{$1818\pm 51 $}  & NA  \\
       &                  &  $M^{\mathrm{2,3}}_{\mathrm{87-87}}$                  &  86.94      &84.49      & 8.3   & $2006\pm 20$ &1.10x \\
       &                  &    
            $M^{\mathrm{2,3}}_{\mathrm{83-83}}$                          & 85.20      & 82.34      & 20.7  & $2328\pm 34$ & 1.28x
                        \\ \cmidrule(l){2-8}
& \multirow{3}{*}{test} & \textbf{baseline}                                                                                      & \textbf{87.64}      & \textbf{85.65}         & \textbf{NA}  &    $1860\pm 36$ &NA  \\
  &                       &   $M^{\mathrm{2,3}}_{\mathrm{87-87}}$              & 86.11      &83.63       & 8.7    &    $2006\pm 83$ & 1.08x         \\
   &                      &         $M^{\mathrm{2,3}}_{\mathrm{83-83}}$                                                       & 84.79      & 82.21      & 20.7   &  $2291\pm 39$ & 1.23x
                            \\ \bottomrule
\multirow{7}{*}{\makecell{MaltParser \\ arc-eager}} & \multirow{3}{*}{dev}  & \textbf{baseline}                                                                                              & \textbf{85.07}      & \textbf{82.65}      & \textbf{NA}  & \textbf{$17387\pm 711$} & NA   \\
 &                        & $M^{\mathrm{2,3}}_{\mathrm{87-87}}$  
                 & 84.38       & 81.64      & 8.3   & \textbf{$18118\pm 860$} & 1.04x\\
&                          
                         &  $M^{\mathrm{2,3}}_{\mathrm{83-83}}$                                      								            & 82.95      & 80.03      & 20.7  & $19758\pm 588$ & 1.14x \\ \cmidrule(l){2-8}
& \multirow{3}{*}{test} & \textbf{baseline}                                                                                             & \textbf{84.58}      & \textbf{82.00}         & \textbf{NA}  &   $17748\pm 801$   & NA     \\
 &                        &  $M^{\mathrm{2,3}}_{\mathrm{87-87}}$               & 83.32      & 80.44      & 8.7    &     $18626\pm 459$ & 1.05x     \\
  &                       &   
                    $M^{\mathrm{2,3}}_{\mathrm{83-83}}$                & 82.02      & 79.13      & 20.7   &     $19286\pm 889$ & 1.09x
                             \\ \midrule
\multirow{7}{*}{\makecell{MaltParser \\ Covington}} & \multirow{3}{*}{dev}  &                                                                            \textbf{baseline}   &  \textbf{83.99}       & \textbf{81.65}      & \textbf{NA}  & \textbf{$16121\pm 581$} & NA   \\
      &                   & $M^{\mathrm{2,3}}_{\mathrm{87-87}}$               & 82.84       & 80.24      & 8.3   &  $16500\pm 819$ & 1.02x\\
       &                  &         
              $M^{\mathrm{2,3}}_{\mathrm{83-83}}$                       & 81.69      & 78.92      & 20.7  & $18210\pm 484$
          &1.13x                \\ \cmidrule(l){2-8}
& \multirow{3}{*}{test} & \textbf{baseline}                                                                                        & \textbf{83.68}      & \textbf{81.34}         & \textbf{NA}  &   $16009\pm 1035$       & NA              \\
  &                       & $M^{\mathrm{2,3}}_{\mathrm{87-87}}$              & 82.75      & 80.02      & 8.7    &    $16561\pm 629$         & 1.03x \\
   &                      &    
                   $M^{\mathrm{2,3}}_{\mathrm{83-83}}$                 & 81.36      & 78.6      & 20.7   &  $17395\pm 1407$ & 1.09x  
                            \\ \midrule

\end{tabular}%
\end{adjustbox}

\label{tab:merged}
\end{table*}

\section{Ongoing Work} \label{sec:work}
One of the main weaknesses of the approach presented above is that the final templates generated during training are only based on adjacent words (i.e., words whose position indexes differ by 1). It does not take into account longer dependency arcs that could be captured by a bigram or trigram after removing the intervening dependents with shorter arcs. This approach can be improved by iteratively finding new templates and recalculating their confidence based on the outcome of applying the preceding template and removing the dependents it captured. In this way, an n-gram would capture a longer arc after intervening words have been removed. In this new approach, the order of applying templates generated during training time is crucial, instead of treating templates as a \emph{bag-of-rules} that match an n-gram from the input.

We performed a preliminary experiment where new templates are generated based on the outcome of applying preceding templates and removing dependents.  We believe it can be beneficial to localize noun phrases first in the input sentence, because they cover vast part of sentences. We look at the distance between PoS tags in an n-gram where the head is a noun. Some PoS tags show tendency to appear closer to a noun than others. We give priority to templates that capture PoS tags closest connected to a noun. In subsequent steps, we generate templates that show the highest confidence at each iteration, and add them to a list that is applied in order. This technique applied on MaltParser with the arc-eager transition system reduces the dev set by almost 21\% with UAS of 82.14 and a LAS of 79.20. 

\section{Conclusion}

We have obtained promising results where the input length can be reduced by more than 20\% at the cost of less than 3 points of UAS. We believe that our work can be a starting point for developing templates that in the future can significantly speed up parsing time by avoiding redundant syntactic analyses at the minimal expense of accuracy. 

The present study has investigated two approaches. In the first technique, which is the main focus of the paper (\Cref{sec:model}) templates were treated as a \emph{bag-of-rules} that have to exceed predefined thresholds for the dominant head and label pattern for a given PoS tag n-gram, prioritizing ones with the highest confidence. In the second approach (\Cref{sec:work}) more importance is given to the order in which templates should be applied. However, the second technique is still in a preliminary stage, and requires some refinement. Research into solving this problem is already in progress. To further our research we plan to use both PoS tags and lemmas in our templates. The sparsity problem in finding n-grams involving lemmas will be tackled by augmenting training data with parsed sentences. 

While we have tested our approach on transition-based dependency parsers, it is worth noting that the technique is generic enough to be applied to practically any kind of parser. Since fragment reuse is implemented as pre and postprocessing step, it works regardless of the inner working of the parser. As the technique reduces the input length received by the parser, speed gains can be expected to be larger on parsers with higher polynomial complexity, like those based on dynamic programming. The same idea would also be applicable to other grammatical representations, for example in constituent parsing, by changing the reusable fragments to the relevant representation (e.g. subtrees of a constituent tree).

Further studies will need to be undertaken in order to show 
the results of the approach when applied on other kinds of parsers, and on other languages different from English.

\section*{Acknowledgments}
This work has received funding from the European
Research Council (ERC), under the European
Union's Horizon 2020 research and innovation
programme (FASTPARSE, grant agreement No
714150), from the TELEPARES-UDC project
(FFI2014-51978-C2-2-R) and the ANSWER-ASAP project (TIN2017-85160-C2-1-R) from MINECO, and from Xunta de Galicia (ED431B 2017/01). We gratefully acknowledge NVIDIA Corporation for the donation of a GTX Titan X GPU.
\clearpage
\bibliographystyle{splncs}
\bibliography{paper}
\end{document}